\documentclass[11pt,a4paper]{article}
\usepackage[hyperref]{naaclhlt2019}
\usepackage{times}
\usepackage{latexsym}
\usepackage{microtype}
\usepackage{booktabs}
\usepackage{url}
\usepackage{eurosym}
\usepackage{todonotes}
\usepackage{multirow}
\usepackage{subcaption}
\usepackage{amssymb}
\usepackage{amsmath}
\usepackage{array}
\aclfinalcopy 

\newcommand{\figref}[1]{Figure~\ref{fig:#1}}
\newcommand{\secref}[1]{Section~\ref{sec:#1}}
\newcommand{\shortsecref}[1]{\S\ref{sec:#1}}
\newcommand{\tabref}[1]{Table~\ref{tab:#1}}
\newcommand{\drop}[0]{DROP}
\newcommand{\squad}[0]{SQuAD}

\newcommand{\PreserveBackslash}[1]{\let\temp=\\#1\let\\=\temp}
\newcolumntype{C}[1]{>{\PreserveBackslash\centering}p{#1}}
\newcolumntype{R}[1]{>{\PreserveBackslash\raggedleft}p{#1}}
\newcolumntype{L}[1]{>{\PreserveBackslash\raggedright}p{#1}}

\title{DROP: A Reading Comprehension Benchmark \\ Requiring
Discrete Reasoning Over Paragraphs}

\author{Dheeru Dua\textsuperscript{$\clubsuit$}, Yizhong Wang\textsuperscript{$\diamondsuit$}\thanks{~Work done as an intern at the Allen Institute for Artificial Intelligence in Irvine, California.}, Pradeep Dasigi\textsuperscript{$\heartsuit$}, \\ \textbf{Gabriel Stanovsky}\textsuperscript{$\heartsuit$}\textsuperscript{$+$}, \textbf{Sameer Singh}\textsuperscript{$\clubsuit$}, and \textbf{Matt Gardner}\textsuperscript{$\spadesuit$} \\
  \textsuperscript{$\clubsuit$}University of California, Irvine, USA \\
  \textsuperscript{$\diamondsuit$}Peking University, Beijing, China \\
  \textsuperscript{$\heartsuit$}Allen Institute for Artificial Intelligence, Seattle, Washington, USA \\
  \textsuperscript{$\spadesuit$}Allen Institute for Artificial Intelligence, Irvine, California, USA \\
  \textsuperscript{$+$}University of Washington, Seattle, Washington, USA \\
  {\tt ddua@uci.edu} \\}

\date{}

\begin{document}
\maketitle
\begin{abstract}

  Reading comprehension has recently seen rapid progress, with systems matching humans on the most popular datasets for the task.  However, a large body of work has highlighted the brittleness of these systems, showing that there is much work left to be done.  We introduce a new English reading comprehension benchmark, DROP, which requires {\bf D}iscrete {\bf R}easoning {\bf O}ver the content of {\bf P}aragraphs.  In this crowdsourced, adversarially-created, 96k-question benchmark, a system must resolve references in a question, perhaps to multiple input positions, and perform discrete operations over them (such as addition, counting, or sorting).  These operations require a much more comprehensive understanding of the content of paragraphs than what was necessary for prior datasets.  We apply state-of-the-art methods from both the reading comprehension and semantic parsing literatures on this dataset and show that the best systems only achieve 32.7\% $F_1$ on our generalized accuracy metric, while expert human performance is 96.4\%.  We additionally present a new model that combines reading comprehension methods with simple numerical reasoning to achieve 47.0\% $F_1$.

\end{abstract}

\section{Introduction}
\label{sec:intro}
The task of \emph{reading comprehension}, where systems must understand a single passage of text well enough to answer arbitrary questions about it, has seen significant progress in the last few years, so much that the most popular datasets available for this task have been solved~\citep{Chen2016ATE,Devlin2018BERTPO}.  We introduce a substantially more challenging English reading comprehension dataset aimed at pushing the field towards more comprehensive analysis of paragraphs of text.  In this new benchmark, which we call DROP, a system is given a paragraph and a question and must perform some kind of {\bf D}iscrete {\bf R}easoning {\bf O}ver the text in the {\bf P}aragraph to obtain the correct answer.

These questions that require discrete reasoning (such as addition, sorting, or counting; see \tabref{main_examples}) are inspired by the complex, compositional questions commonly found in the semantic parsing literature.  We focus on this type of questions because they force a structured analysis of the content of the paragraph that is detailed enough to permit reasoning.  Our goal is to further \emph{paragraph understanding}; complex questions allow us to test a system's understanding of the paragraph's semantics.

DROP is also designed to further research on methods that combine distributed representations with symbolic, discrete reasoning.  In order to do well on this dataset, a system must be able to find multiple occurrences of an event described in a question (presumably using some kind of soft matching), extract arguments from the events, then perform a numerical operation such as a sort, to answer a question like \textit{``Who threw the longest touchdown pass?''}.

We constructed this dataset through crowdsourcing, first collecting passages from Wikipedia that are easy to ask hard questions about, then encouraging crowd workers to produce challenging questions.  This encouragement was partially through instructions given to workers, and partially through the use of an \emph{adversarial baseline}: we ran a baseline reading comprehension method (BiDAF)~\citep{Seo2016BidirectionalAF} in the background as crowd workers were writing questions, requiring them to give questions that the baseline system could not correctly answer.  This resulted in a dataset of 96,567 questions from a variety of categories in Wikipedia, with a particular emphasis on sports game summaries and history passages.  The answers to the questions are required to be spans in the passage or question, numbers, or dates, which allows for easy and accurate evaluation metrics.

\begin{table*}[!t]
\centering
\footnotesize
\begin{tabular}{L{1.4cm}p{7cm}p{2.8cm}L{1.25cm}L{1.35cm}}
\toprule
{\bf Reasoning} & {\bf Passage} (some parts shortened) & {\bf Question} & {\bf Answer} & {\bf BiDAF}\\
 \midrule
 Subtraction (28.8\%) & That year, his {\bf \color{teal} Untitled (1981)}, a painting of a haloed, black-headed man with a bright red skeletal body, depicted amid the artists signature scrawls, was {\bf \color{teal}{sold by Robert Lehrman for \$16.3 million, well above its  \$12 million high estimate}}.  & How many more dollars was the Untitled (1981) painting sold for than the 12 million dollar estimation? & 4300000 & \$16.3 million \\ 
 \midrule
 Comparison (18.2\%) & In {\bf \color{orange}1517, the seventeen-year-old King sailed to Castile}. There, his Flemish court \ldots. {\bf \color{orange}In May 1518, Charles traveled to Barcelona in Aragon}. & Where did Charles travel to first, Castile or Barcelona? & Castile & Aragon\\
 \midrule
 Selection (19.4\%) & In 1970, to commemorate the 100th anniversary of the founding of Baldwin City, {\bf \color{purple}Baker University professor and playwright Don Mueller and Phyllis E. Braun, Business Manager, produced a musical play entitled The Ballad Of Black Jack} to tell the story of the events that led up to the battle. & Who was the University professor that helped produce The Ballad Of Black Jack, Ivan Boyd or Don Mueller? & Don Mueller & Baker\\
 \midrule
 Addition (11.7\%) & Before the UNPROFOR fully deployed, the HV clashed with an armed force of the RSK in the village of Nos Kalik, located in a pink zone near \v{S}ibenik, and captured the village at 4:45 p.m. on {\bf \color{red} 2 March 1992}. The JNA formed a battlegroup to counterattack the {\bf \color{red} next day}. & What date did the JNA form a battlegroup to counterattack after the village of Nos Kalik was captured? & 3 March 1992 & 2 March 1992 \\ 
 \midrule
 Count (16.5\%) and Sort (11.7\%) & Denver would retake the lead with kicker {\bf \color{brown}Matt Prater nailing a 43-yard field goal}, yet Carolina answered as kicker {\bf \color{brown}John Kasay ties the game with a 39-yard field goal}. \ldots \ Carolina closed out the half with {\bf \color{brown}Kasay nailing a 44-yard field goal}. \ldots \ In the fourth quarter, Carolina sealed the win with {\bf \color{brown}Kasay's 42-yard field goal}. & Which kicker kicked the most field goals? & John Kasay & Matt Prater\\
 \midrule
 Coreference Resolution (3.7\%) & \ {\bf \color{violet}James Douglas} was the second son of Sir George Douglas of Pittendreich, and Elizabeth Douglas, daughter David Douglas of Pittendreich. Before {\bf \color{violet}1543 he married Elizabeth}, daughter of James Douglas, 3rd Earl of Morton. {\bf \color{violet}In 1553 James Douglas succeeded to the title and estates of his father-in-law}. & How many years after he married Elizabeth did James Douglas succeed to the title and estates of his father-in-law? & 10 & 1553\\
 \midrule
 
Other Arithmetic (3.2\%) & Although the movement initially gathered some {\bf \color{magenta}60,000 adherents}, the subsequent establishment of the Bulgarian Exarchate {\bf \color{magenta}reduced their number by some 75\%}. & How many adherents were left after the establishment of the Bulgarian Exarchate? & 15000 & 60,000\\
\midrule
Set of spans (6.0\%) &  According to some sources 363 civilians were killed in {\bf \color{blue}Kavadarci}, 230 in {\bf \color{blue}Negotino} and 40 in {\bf \color{blue}Vatasha}. & What were the 3 villages that people were killed in? & Kavadarci, Negotino, Vatasha & Negotino and 40 in Vatasha \\
\midrule
Other (6.8\%) & This {\bf \color{olive}Annual Financial Report} is our principal financial statement of accountability. The {\bf \color{olive}AFR gives a comprehensive view} of the Department's financial activities ... & What does AFR stand for? & Annual Financial Report & one of the Big Four audit firms\\
 \bottomrule
\end{tabular}
\caption{Example questions and answers from the \drop~dataset, showing the relevant parts of the associated passage and the reasoning required to answer the question.}
\label{tab:main_examples}
\end{table*}

We present an analysis of the resulting dataset to show what phenomena are present.  We find that many questions combine complex question semantics with SQuAD-style argument finding; e.g., in the first question in \tabref{main_examples}, BiDAF correctly finds the amount the painting sold for, but does not understand the question semantics and cannot perform the numerical reasoning required to answer the question.  Other questions, such as the fifth question in \tabref{main_examples}, require finding all events in the passage that match a description in the question, then aggregating them somehow (in this instance, by counting them and then performing an argmax).  Very often entity coreference is required.  \tabref{main_examples} gives a number of different phenomena, with their proportions in the dataset. 

We used three types of systems to judge baseline performance on DROP: (1) heuristic baselines, to check for biases in the data; (2) SQuAD-style reading comprehension methods; and (3) semantic parsers operating on a pipelined analysis of the passage.  The reading comprehension methods perform the best, with our best baseline achieving 32.7\% $F_1$ on our generalized accuracy metric, while expert human performance is 96.4\%.
Finally, we contribute a new model for this task that combines limited numerical reasoning with standard reading comprehension methods, allowing the model to answer questions involving counting, addition and subtraction.  This model reaches 47\% $F_1$, a 14.3\% absolute increase over the best baseline system.

The dataset, code for the baseline systems, and a leaderboard with a hidden test set can be found at \url{https://allennlp.org/drop}.

\section{Related Work}
\label{sec:related_work}
\textbf{Question answering datasets} With systems reaching human performance on the Stanford Question Answering Dataset (SQuAD) \citep{Rajpurkar2016SQuAD10}, many follow-on tasks are currently being proposed.  All of these datasets throw in additional complexities to the reading comprehension challenge, around tracking conversational state~\citep{Reddy2018CoQAAC,Choi2018QuACQA}, requiring passage retrieval~\citep{Joshi2017TriviaQAAL,Yang2018HotpotQAAD,Talmor2018TheWA}, mismatched passages and questions~\citep{Saha2018DuoRCTC,Kocisk2018TheNR,Rajpurkar2018KnowWY}, integrating knowledge from external sources~\citep{Mihaylov2018CanAS,Zhang2018ReCoRDBT}, tracking entity state changes~\citep{mishra2018tracking,ostermann2018mcscript} or a particular kind of ``multi-step'' reasoning over multiple documents~\citep{Welbl2018ConstructingDF,Khashabi2018LookingBT}. Similar facets are explored in medical domain datasets \citep{pampari2018emrqa,vsuster2018clicr} which contain automatically generated queries on medical records based on predefined templates. We applaud these efforts, which offer good avenues to study these additional phenomena.  However, we are concerned with \emph{paragraph understanding}, which on its own is far from solved, so DROP has none of these additional complexities.  It consists of single passages of text paired with independent questions, with only linguistic facility required to answer the questions.\footnote{Some questions in our dataset require limited sports domain knowledge to answer; we expect that there are enough such questions that systems can reasonably learn this knowledge from the data.}  One could argue that we are adding numerical reasoning as an ``additional complexity'', and this is true; however, it is only simple reasoning that is relatively well-understood in the semantic parsing literature, and we use it as a necessary means to force more comprehensive passage understanding.

Many existing algebra word problem datasets also contain similar phenomena to what is in DROP~\citep{KoncelKedziorski2015ParsingAW,kushman2014learning,hosseini2014learning,clark2016combining,Ling2017ProgramIB}. Our dataset is different in that it uses much longer contexts, is more open domain, and requires deeper paragraph understanding.

\textbf{Semantic parsing} The semantic parsing literature has a long history of trying to understand complex, compositional question semantics in terms of some grounded knowledge base or other environment~\citep[\emph{inter alia}]{Zelle1996LearningTP,Zettlemoyer2005LearningTM,Berant2013SemanticPO}.  It is this literature that we modeled our questions on, particularly looking at the questions in the WikiTableQuestions dataset~\citep{Pasupat2015CompositionalSP}.  If we had a structured, tabular representation of the content of our paragraphs, DROP would be largely the same as WikiTableQuestions, with similar (possibly even simpler) question semantics.  Our novelty is that we are the first to combine these complex questions with paragraph understanding, with the aim of encouraging systems that can produce comprehensive structural analyses of paragraphs, either explicitly or implicitly.

\textbf{Adversarial dataset construction} We continue a recent trend in creating datasets with adversarial baselines in the loop~\citep{Paperno2016TheLD,Minervini2018AdversariallyRN,Zellers2018SWAGAL,Zhang2018ReCoRDBT,zellers2018VCR}.  In our case, instead of using an adversarial baseline to filter automatically generated examples, we use it in a crowdsourcing task, to teach crowd workers to avoid easy questions, raising the difficulty level of the questions they provide.

\textbf{Neural symbolic reasoning} DROP is designed to encourage research on methods that combine neural methods with discrete, symbolic reasoning.  We present one such model in \secref{model}.  Other related work along these lines has been done by \citet{Reed2015NeuralP}, \citet{Neelakantan2015NeuralPI}, and \citet{Liang2017NeuralSM}.

\section{DROP Data Collection}
\label{sec:data_collection}
In this section, we describe our annotation protocol, which consists of three phases. First, we automatically extract passages from Wikipedia which are expected to be amenable to complex questions.
Second, we crowdsource question-answer pairs on these passages, eliciting questions which require discrete reasoning. Finally, we validate the development and test portions of \drop~to ensure their quality and report inter-annotator agreement.

\paragraph{Passage extraction} We searched Wikipedia for passages that had a narrative sequence of events, particularly with a high proportion of numbers, as our initial pilots indicated that these passages were the easiest to ask complex questions about.  We found that National Football League (NFL) game summaries and history articles were particularly promising, and we additionally sampled from any Wikipedia passage that contained at least twenty numbers.\footnote{We used an October 2018 Wikipedia dump, as well as scraping of online Wikipedia.} This process yielded a collection of about 7,000 passages.


\paragraph{Question collection} 
We used Amazon Mechanical Turk\footnote{\url{www.mturk.com}} to crowdsource the collection of question-answer pairs, 
where each question could be answered in the context of a single Wikipedia passage.
In order to allow some flexibility during the annotation process, 
in each human intelligence task (HIT) workers were presented with a random sample of 5 of our Wikipedia passages, 
and were asked to produce a total of at least 12 question-answer pairs on any of these.

We presented workers with example questions from five main categories, inspired by questions from the semantic parsing literature (addition/subtraction, minimum/maximum, counting, selection and comparison; see examples in \tabref{main_examples}), to elicit questions that require complex linguistic understanding and discrete reasoning. 
In addition, to further increase the difficulty of the questions in \drop, we employed a novel adverserial annotation setting, 
where workers were only allowed to submit questions which a real-time QA model BiDAF \emph{could not} solve.\footnote{While BiDAF is no longer state-of-the-art, performance is reasonable and the AllenNLP implementation \cite{Gardner2017AllenNLP} made it the easiest to deploy as a server.}

Next, each worker answered their own question with one of three answer types: spans of text from either question or passage, a date (which was common in history and open-domain text) and numbers, allowed only for questions which explicitly stated a specific unit of measurement (e.g., ``How many \emph{yards} did Brady run?''), in an attempt to simplify the evaluation process. 

Initially, we opened our HITs to all United States workers and gradually reduced our worker pool to workers who understood the task and annotated it well. Each HIT paid 5 USD and could be completed within 30 minutes, compensating a trained worker with an average pay of 10 USD/ hour. 

Overall, we collected a total of 96,567 question-answer pairs with a total Mechanical Turk budget of 60k USD (including validation). The dataset was randomly partitioned by passage into training (80\%), development (10\%) and test (10\%) sets, so all questions about a particular passage belong to only one of the splits.

\paragraph{Validation}
In order to test inter-annotator agreement and to improve the quality of evaluation against \drop, 
we collected at least two additional answers for each question in the development and test sets. 

In a separate HIT, workers were given context passages and a previously crowdsourced question, and were asked to either answer the question or mark it as invalid (this occurred for 0.7\% of the data, which we subsequently filtered out).
We found that the resulting inter-annotator agreement was good and on par with other QA tasks; overall Cohen's $\kappa$ was 0.74, with 0.81 for numbers, 0.62 for spans, and 0.65 for dates. 



\section{DROP Data Analysis}
\label{sec:data_analysis}

  \begin{table}[t]
\centering
\small
\begin{tabular}{@{}lrrrr@{}}
\toprule
{\bf Statistic}         & {\bf Train} & {\bf Dev} & {\bf Test}  \\ \midrule
Number of passages        & 5565          & 582     & 588       \\
Avg. passage len [words]  & 213.45      & 191.62      & 195.12        \\
Number of questions       & 77,409        & 9,536      & 9,622        \\
Avg. question len [words]  & 10.79     & 11.17           & 11.23        \\
Avg. questions / passage  & 13.91        & 16.38        & 16.36        \\
Question vocabulary size   & 29,929      & 8,023       & 8,007        \\  \bottomrule
\end{tabular}
\caption{Dataset statistics across the different splits.}
    \label{tab:dataset_stats}    
\end{table}

In the following, we quantitatively analyze properties of passages, questions, and answers in \drop.
Different statistics of the dataset are depicted in \tabref{dataset_stats}. Notably, questions have a diverse vocabulary of around 30k different words in our training set.

\begin{figure*}[tb]
    \centering
    \begin{subfigure}[t]{0.49\textwidth}
        \centering
        \includegraphics[clip,trim=100 25 100 15,width=\textwidth]{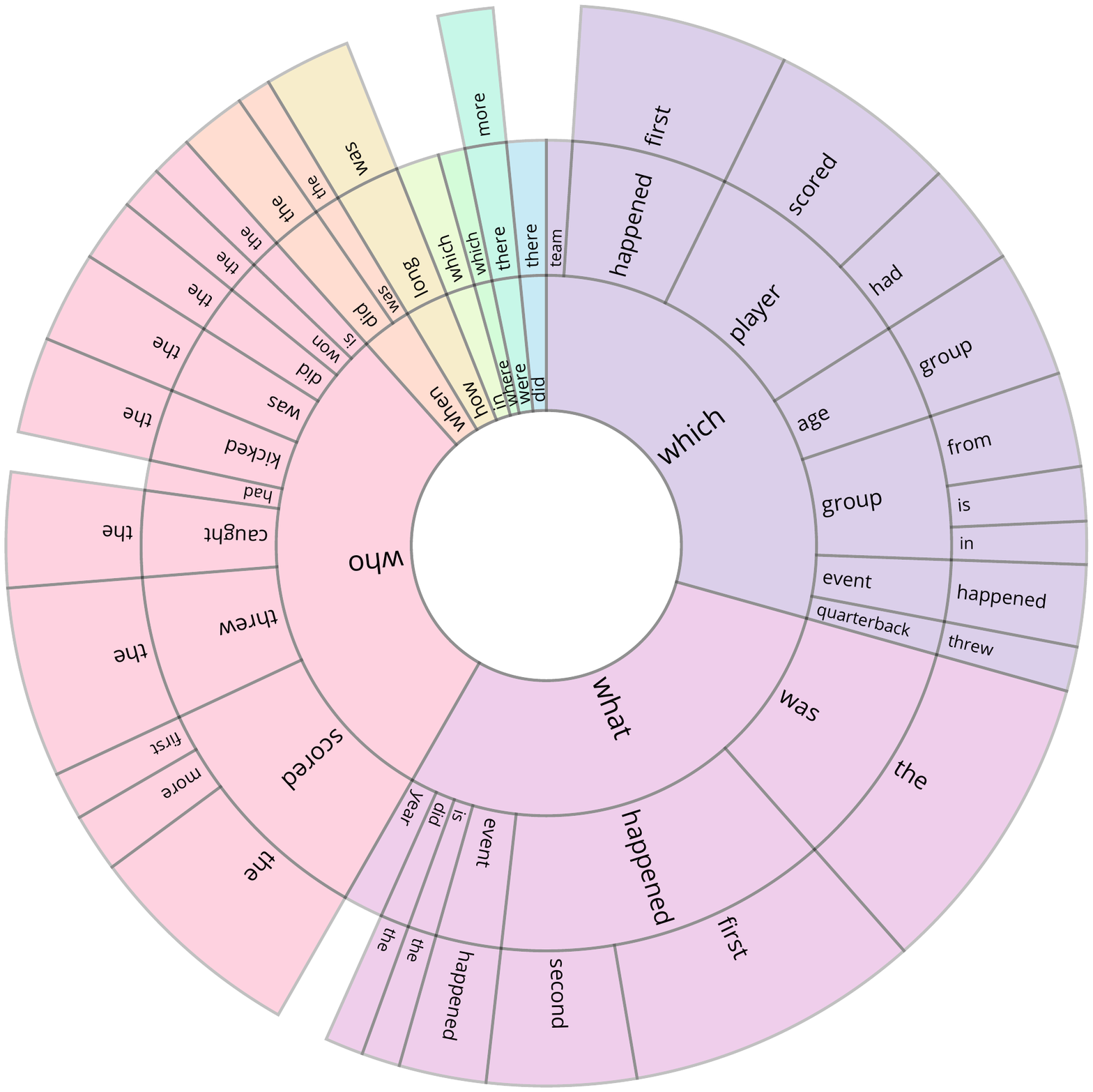}
        \caption{For span type answers}
        \label{fig:question_patterns_span}
    \end{subfigure}
    \begin{subfigure}[t]{0.49\textwidth}
        \centering
        \includegraphics[clip,trim=100 25 100 15,width=\textwidth]{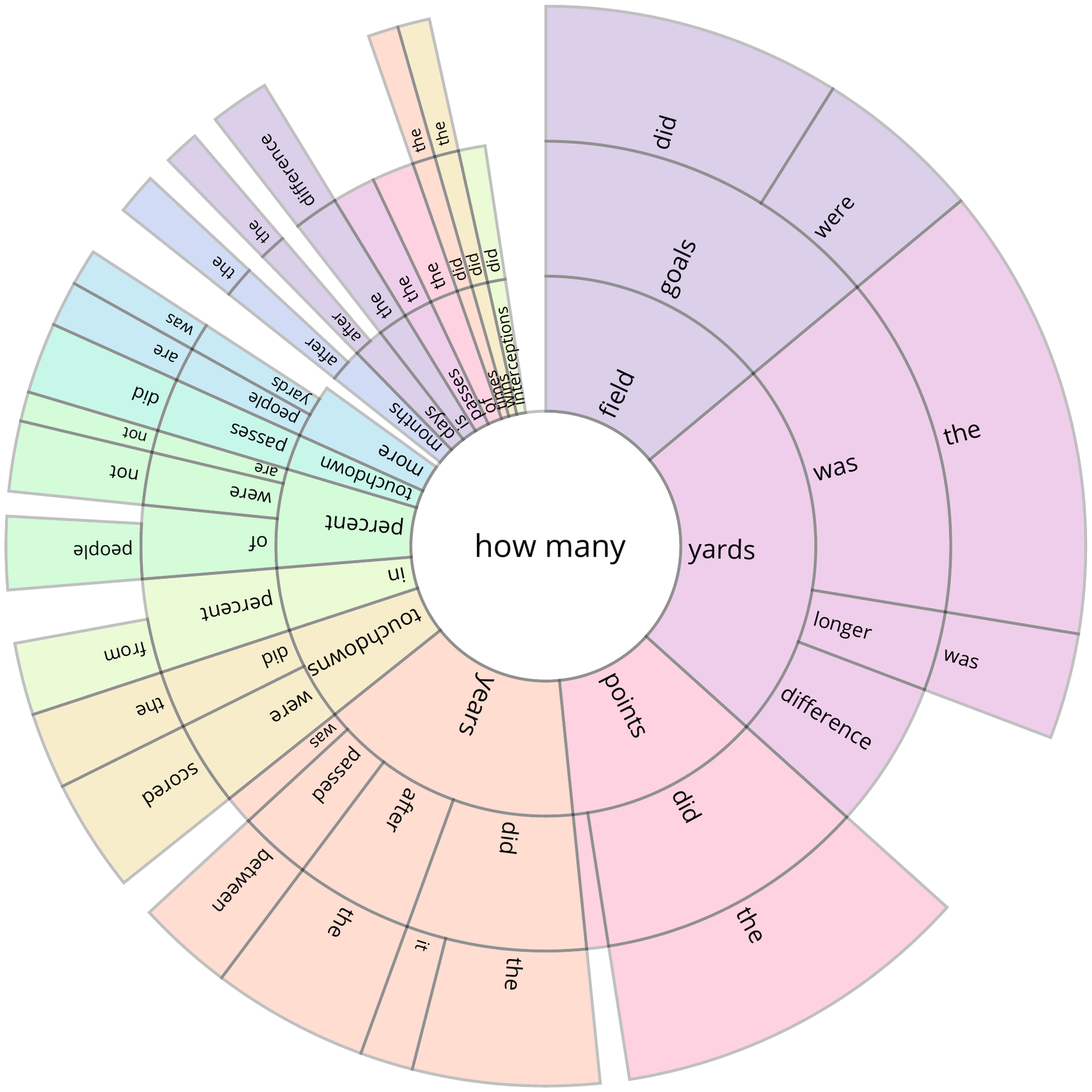}
        \caption{For number type answers}
        \label{fig:question_patterns_number}
    \end{subfigure}
    \caption{Distribution of the most popular question prefixes for two different subsets of the training data.}
    \label{fig:question_patterns}
\end{figure*}

\paragraph{Question analysis}
To assess the question type distribution,
we sampled 350 questions from the training and development sets and manually annotated the categories of discrete operations required to answer the question. \tabref{main_examples} shows the distribution of these categories in the dataset.
In addition, to get a better sense of the lexical diversity of questions in the dataset, 
we find the most frequent trigram patterns in the questions per answer type.  
We find that the dataset offers a huge variety of linguistic constructs, with the most frequent pattern (``Which team scored") appearing  
in only 4\%  of the span type questions. 
For number type questions, the 5 most frequent question patterns all start with ``How many", indicating the need to perform counting and other arithmetic operations. A distribution of the trigrams containing the start of the questions are shown in \figref{question_patterns}.

\paragraph{Answer analysis}
To discern the level of \emph{passage understanding} needed to answer the questions in \drop, we annotate the set of spans in the passage that are necessary for answering the 350 questions mentioned above. We find that on an average 2.18 spans need to be considered to answer a question and the average distance between these spans is 26 words, with 20\% of samples needing at least 3 spans (see appendix for examples).
Finally, we assess the answer distribution in \tabref{answer_types}, by running the part-of-speech tagger and named entity recognizer from spaCy\footnote{https://spacy.io/} to automatically partition all the answers into various categories. We find that a majority of the answers are numerical values and proper nouns.




    

  \begin{table}[t]
  \centering
  \small
    \begin{tabular}{lrr}
    \toprule
    {\bf Answer Type} & {\bf Percent} & {\bf Example}\\
     \midrule
     NUMBER & 66.1 & 12 \\
     PERSON & 12.2 & Jerry Porter \\ 
     OTHER & 9.4 & males\\
     OTHER ENTITIES & 7.3 & Seahawks \\ 
     VERB PHRASE & 3.5 & Tom arrived at Acre \\ 
     DATE & 1.5 & 3 March 1992 \\ 
     \bottomrule
    \end{tabular}
    \caption{Distribution of answer types in training set, according to an automatic named entity recognition.}
    \label{tab:answer_types}
  \end{table}

\section{Baseline Systems}
\label{sec:baselines}
In this section we describe the initial baselines that we evaluated on the
\drop~ dataset. We used three types of baselines:
state-of-the-art semantic parsers (\shortsecref{semparse}), state-of-the-art reading comprehension models (\shortsecref{rc}), and heuristics looking for annotation artifacts (\shortsecref{heuristics}).  We use two evaluation metrics to compare model performance: \emph{Exact-Match}, and a numeracy-focused (macro-averaged) \emph{$F_1$} score, which measures overlap between a bag-of-words representation of the gold and predicted answers. We employ the same implementation of Exact-Match accuracy as used by \squad{}, which removes articles and does other simple normalization, and our $F_1$ score is based on that used by \squad{}. Since \drop{} is numeracy-focused, we define $F_1$ to be 0 when there is a number mismatch between the gold and predicted answers, regardless of other word overlap. When an answer has multiple spans, we first perform a one-to-one alignment greedily based on bag-of-word overlap on the set of spans and then compute average $F_1$ over each span. When there are multiple annotated answers, both metrics take a max over all gold answers.

\subsection{Semantic Parsing}
\label{sec:semparse}

Semantic parsing
has been used to translate natural language utterances into formal executable languages (e.g., SQL) that can perform discrete operations
against a structured knowledge representation, such as knowledge graphs or tabular databases ~\citep[\emph{inter alia}]{Zettlemoyer2005LearningTM,berant2013semantic,Yin2017ASN,chen2011learning}.
Since many of \drop's questions require similar discrete reasoning, it is appealing
to port some of the successful work in semantic parsing to the \drop~ dataset.
Specifically, we use the grammar-constrained semantic parsing model
built by~\citet{Krishnamurthy2017neuralsp} (KDG) for the \textsc{WikiTableQuestions} tabular dataset~\citep{Pasupat2015CompositionalSP}. 

\paragraph{Sentence representation schemes
}
We experimented with three paradigms to represent paragraphs as structured contexts:
(1) Stanford dependencies \cite[Syn Dep]{Marneffe2008TheST}; which capture word-level syntactic relations,
(2) Open Information Extraction \citep[Open IE]{Banko2007OpenIE}, a
shallow semantic representation which directly links predicates
and arguments; 
and
(3) Semantic Role Labeling \citep[SRL]{SRL}, which disambiguates senses for polysemous predicates and assigns predicate-specific
argument roles.\footnote{We used the AllenNLP implementations of state-of-the-art models for all of these representations \cite{Gardner2017AllenNLP,Dozat2017StanfordsGN,He2017DeepSR,Stanovsky2018SupervisedOI}.}
To adhere to KDG's structured representation format,
we convert each of these representations into a table, where rows are predicate-argument structures and columns correspond to different argument roles.


\paragraph{Logical form language}
Our logical form language identifies five basic elements in the table representation: \textit{predicate-argument structures} (i.e., table rows), \textit{relations} (column-headers), \textit{strings}, \textit{numbers}, and \textit{dates}. 
In addition, it defines functions that operate on these elements,
such as counters and filters.\footnote{For example \texttt{filter\_number\_greater} takes a set of predicate-argument structures, the name of a relation, and a number, and returns all those structures where the numbers in the argument specified by the relation are greater than the given number.} 
Following~\citet{Krishnamurthy2017neuralsp}, we use the argument and return types of these functions to automatically induce a grammar to constrain the parser.
We also add context-specific rules to produce strings occurring in both question and paragraph, and those paragraph strings that are neighbors of question tokens in the GloVe embedding space \cite{Pennington2014GloveGV}, up to a cosine distance of $d$.\footnote{$d=0.3$ was manually tuned on the development set.}
The complete set of functions used in our language and their induced grammar can be found in the code release. 

\paragraph{Training and inference}
During training, the KDG parser maximizes the marginal likelihood of a set of (possibly spurious) question logical forms that evaluate to the correct answer. We obtain this set by performing an exhaustive search over the grammar up to a preset tree depth. At test time, we use beam search to produce the most likely logical form, which is then executed to predict an answer.


\subsection{SQuAD-style Reading Comprehension}
\label{sec:rc}


We test four different SQuAD-style reading comprehension models on \drop:
(1) \textbf{BiDAF}  \cite{Seo2016BidirectionalAF}, which is the adversarial baseline we used in data construction (66.8\% EM on \squad~1.1); 
(2) \textbf{QANet} \cite{yu2018qanet}, currently the best-performing published model on SQuAD 1.1 without data augmentation or pre-training
(72.7\% EM); 
(3) \textbf{QANet + ELMo}, which enhances the QANet model by concatenating pre-trained ELMo representations \cite{peters2018elmo} to the original embeddings
(78.7\% EM);
(4) \textbf{BERT} \cite{Devlin2018BERTPO}, which recently achieved improvements on many NLP tasks with a novel pre-training technique (84.7\% EM).\footnote{The first three scores are based on our own implementation, while the score for BERT is based on an open-source implementation from Hugging Face: https://github.com/huggingface/pytorch-pretrained-bert}



These models require a few minor adaptations when training on \drop.
While SQuAD provides answer indices in the passage, our dataset only provides the answer strings. To address this, we use the marginal likelihood objective function proposed by \citet{Clark2018SimpleAE}, which sums over the probabilities of all the matching spans.\footnote{For the black-box BERT model, we convert \drop~to \squad~format by using the first match as the gold span.}
We also omitted the training questions which cannot be answered by a span in the passage (45\%), and therefore cannot be represented by these systems.

For the BiDAF baseline, we use the implementation in AllenNLP but change it to use the marginal objective. For the QANet model, our settings differ from the original paper only in the batch size (16 v.s. 32) and number of blocks in the modeling layer (6 v.s. 7) due to the GPU memory limit. We adopt the ELMo representations trained on 5.5B corpus for the QANet+ELMo baseline and the large uncased BERT model for the BERT baseline. The hyper-parameters for our NAQANet model (\S\ref{sec:model}) are the same as for the QANet baseline. 

\subsection{Heuristic Baselines}
\label{sec:heuristics}


A recent line of work \cite{Gururangan:2018,Kaushik2018HowMR} has identified that popular crowdsourced NLP datasets
(such as SQuAD \cite{Rajpurkar2016SQuAD10} or SNLI \cite{Bowman2015ALA}) are prone to have artifacts and annotation biases which can be exploited by supervised algorithms that
learn to pick up these artifacts as signal instead of more meaningful semantic features.
%
We estimate artifacts by training the QANet model described in \secref{rc} on a version of \drop~where either the question or the 
paragraph input representation vectors are zeroed out ({\bf question-only} and {\bf paragraph-only}, respectively).
Consequently, the resulting models can then only predict answer spans from either the question or the paragraph.

In addition, we devise a baseline that estimates the answer variance in \drop.
We start by counting the unigram and bigram answer frequency 
for each wh question-word in the train set (as the first word in the question). 
The {\bf majority baseline} then predicts an answer as the set of 3 most common answer spans for the input question word (e.g., for ``when'', these were ``quarter'', ``end'' and ``October'').

\section{NAQANet}
\label{sec:model}
DROP is designed to encourage models that combine neural reading comprehension with symbolic reasoning.  None of the baselines we described in \secref{baselines} can do this.  As a preliminary attempt toward this goal, we propose a numerically-aware QANet model, NAQANet, which allows the state-of-the-art reading comprehension system to produce three new answer types: (1) spans from the question; (2) counts; (3) addition or subtraction over numbers.  To predict numbers, the model first predicts whether the answer is a count or an arithmetic expression.  It then predicts the specific numbers involved in the expression.  This can be viewed as the neural model producing a \emph{partially executed} logical form, leaving the final arithmetic to a symbolic system.  While this model can currently only handle a very limited set of operations, we believe this is a promising approach to combining neural methods and symbolic reasoning.  The model is trained by marginalizing over all execution paths that lead to the correct answer.

\subsection{Model Description}
Our NAQANet model follows the typical architecture of previous reading comprehension models, which is composed of embedding, encoding, passage-question attention, and output layers.  We use the original QANet architecture for everything up to the output layer. This gives us a question representation $\textbf{Q}\in\mathbb{R}^{m \times d}$, and a projected question-aware passage representation $\bar{\textbf{P}}\in\mathbb{R}^{n \times d}$. We have four different output layers, for the four different kinds of answers the model can produce:

\paragraph{Passage span}
As in the original QANet model, to predict an answer in the passage we apply three repetitions of the QANet encoder to the passage representation $\bar{\textbf{P}}$ and get their outputs as $\textbf{M}_0$, $\textbf{M}_1$, $\textbf{M}_2$ respectively. Then the probabilities of the starting and ending positions from the passage can be computed as:
\begin{align}
    \mathbf{p}^\textrm{p\_start} &= \textrm{softmax}(\textrm{FFN}([\textbf{M}_0; \textbf{M}_1]), \\
    \mathbf{p}^\textrm{p\_end} &= \textrm{softmax}(\textrm{FFN}([\textbf{M}_0; \textbf{M}_2])
\end{align}
where FFN is a two-layer feed-forward network with the RELU activation. 

\paragraph{Question span}
Some questions in DROP have their answer in the \emph{question} instead of the passage. To predict an answer from the question, the model first computes a vector $\mathbf{h}^P$ that represents the information it finds in the passage:
\begin{align}
    \boldsymbol{\alpha}^P &= \textrm{softmax}(\mathbf{W}^P \bar{\textbf{P}}), \\
    \mathbf{h}^P &= \boldsymbol{\alpha}^P \bar{\textbf{P}}
\end{align}
Then it computes the probabilities of the starting and ending positions from the question as:
\begin{align}
    \mathbf{p}^\textrm{q\_start} &= \textrm{softmax}(\textrm{FFN}([\textbf{Q}; \mathbf{e}^{|Q|}\otimes\mathbf{h}^P ]), \\
    \mathbf{p}^\textrm{q\_end} &= \textrm{softmax}(\textrm{FFN}([\textbf{Q}; \mathbf{e}^{|Q|}\otimes\mathbf{h}^P ])
\end{align}
where the outer product with the identity $(\mathbf{e}^{|Q|} \otimes \mathbf{\cdot})$ simply repeats $\mathbf{h}^P$ for each question word.

\paragraph{Count}
We model the capability of counting as a multi-class classification problem. Specifically, we consider ten numbers (0--9) in this preliminary model and the probabilities of choosing these numbers is computed based on the passage vector $\mathbf{h}^P$:
\begin{equation}
    \mathbf{p}^\textrm{count} = \textrm{softmax}(\textrm{FFN}(\mathbf{h}^P))
\end{equation}

\paragraph{Arithmetic expression}
Many questions in DROP require the model to locate multiple numbers in the passage and add or subtract them to get the final answer. To model this process, we first extract all the numbers from the passage and then learn to assign a plus, minus or zero for each number. In this way, we get an arithmetic expression composed of signed numbers, which can be evaluated to give the final answer. 

To do this, we first apply another QANet encoder to $\textbf{M}_2$ and get a new passage representation $\textbf{M}_3$. Then we select an index over the concatenation of $\textbf{M}_0$ and $\textbf{M}_3$, to get a representation for each number in this passage. The $i^{th}$ number can be represented as ${\mathbf{h}_i^{N}}$ and the probabilities of this number being assigned a plus, minus or zero are:
\begin{equation}
    \mathbf{p}_i^\textrm{sign} = \textrm{softmax}(\textrm{FFN}(\mathbf{h}_i^N))
\end{equation}

\paragraph{Answer type prediction}
We use a categorical variable to decide between the above four answer types, with probabilities computed as:
\begin{equation}
    \mathbf{p}^\textrm{type} = \textrm{softmax}(\textrm{FFN}([\mathbf{h}^P, \mathbf{h}^Q]))
\end{equation}
where $\mathbf{h}^Q$ is computed over $\mathbf{Q}$, in a similar way as we did for $\mathbf{h}^P$. At test time, we first determine this answer type greedily and then get the best answer from the selected type. 

\subsection{Weakly-Supervised Training}

For supervision, DROP contains only the answer string, not which of the above answer types is used to arrive at the answer. To train our model, we adopt the weakly supervised training method widely used in the semantic parsing literature \cite{Berant2013SemanticPO}. We find all executions that evaluate to the correct answer, including matching passage spans and question spans, correct count numbers, as well as sign assignments for numbers. Our training objective is then to maximize the marginal likelihood of these executions.\footnote{Due to the exponential search space and the possible noise, we only search the addition/subtraction of two numbers.  Given this limited search space, the search and marginalization are exact.}

\begin{table}
    \small
    \centering
    \begin{tabular}{lrrrr}
    \toprule
     \multirow{2}{*}{\bf Method}    & \multicolumn{2}{c}{\bf Dev} & \multicolumn{2}{c}{\bf Test} \\
     \cmidrule(lr){2-3}
     \cmidrule(lr){4-5}
         & EM & F$_1$ & EM&  F$_1$\\
    \midrule

    \multicolumn{5}{l}{\bf Heuristic Baselines}\\
    Majority   &       0.09    &       1.38    &       0.07    &       1.44   \\
    Q-only     &       4.28   &       8.07   &       4.18    &       8.59   \\
    P-only     &      0.13    &       2.27    &       0.14    &       2.26   \\
    \addlinespace

    \multicolumn{5}{l}{\bf Semantic Parsing}\\
    Syn Dep     &   9.38    &       11.64   &       8.51    &   10.84  \\
    OpenIE      &   8.80    &       11.31   &       8.53    &   10.77  \\
    SRL         &   9.28   &       11.72   &       8.98   &     11.45   \\
    \addlinespace
         
    \multicolumn{5}{l}{\bf SQuAD-style RC}\\
    BiDAF       &       26.06   &       28.85   &       24.75   &       27.49   \\
    QANet       &       27.50   &       30.44   &       25.50   &       28.36   \\
    QANet+ELMo  &       27.71   &       30.33   &       27.08   &       29.67   \\
    BERT        &       30.10   &       33.36   &       29.45   &       32.70   \\
    \addlinespace

    \multicolumn{5}{l}{\bf NAQANet}\\
    + Q Span   &       25.94   &      29.17   &       24.98   &       28.18   \\
    + Count     &       30.09   &   33.92   &       30.04   &       32.75   \\
    + Add/Sub     &       43.07   &      45.71   &       40.40   &       42.96   \\
    Complete Model      &   {\bf 46.20}   &   {\bf 49.24}   &   {\bf 44.07}   &   {\bf 47.01}   \\
    \addlinespace

    \bf Human   &   -   &   -   &    94.09   &   96.42   \\
    \bottomrule
    \end{tabular}
    \caption{Performance of the different models on our development and test set, in terms of Exact Match (EM), and numerically-focused $F_1$ (\S\ref{sec:baselines}). Both metrics are calculated as the maximum against a set of gold answers.}
    \label{tab:results}
\end{table}

\begin{table*}[!t]
\centering
\footnotesize
\begin{tabular}{p{2cm}p{5cm}p{4.4cm}cc} 
\toprule
\multirow{2}{*}{\bf Phenomenon} & \multirow{2}{*}{\bf Passage Highlights} & \multirow{2}{*}{\bf Question} & \multirow{2}{*}{\bf Answer} & {\bf Our}\\
& & & & {\bf model}\\
 \midrule
Subtraction +~~Coreference & 
\ldots{} Twenty-five of his 150 men were sick, and his advance stalled \ldots & 
How many of Bartolomé de Amésqueta's 150 men were not sick? & 
125 & 
145\\
\midrule
Count + Filter & 
\ldots{} Macedonians were the largest ethnic group in Skopje, with 338,358 inhabitants \ldots{} Then came \ldots{} Serbs (14,298 inhabitants), Turks (8,595), Bosniaks (7,585) and Vlachs (2,557) \ldots & 
How many ethnicities had less than 10000 people? & 
3 & 
2\\
\midrule
Domain ~~~~~~~~knowledge & 
\ldots{} Smith was sidelined by a torn pectoral muscle suffered during practice \ldots &
How many quarters did Smith play? &
0 & 
2 \\
 \midrule
Addition & 
\ldots{} culminating in the Battle of Vienna of 1683, which marked the start of the 15-year-long Great Turkish War \ldots & 
What year did the Great Turkish War end? & 
1698 & 
1668\\
 \bottomrule
\end{tabular}
\caption{Representative examples from our model's error analysis. We list the identified semantic phenomenon, the relevant passage highlights, a gold question-answer pair, and the erroneous prediction by our model.}
\label{tab:error_analysis}
\end{table*}

\section{Results and Discussion}
\label{sec:results}




The performance of all tested models on the 
\drop~dataset is presented in \tabref{results}.
Most notably, all models perform significantly worse than on other prominent reading comprehension datasets, while human performance remains at similar high levels.\footnote{Human performance was estimated by the authors collectively answering 560 questions from the test set, which were then evaluated using the same metric as learned systems. This is in contrast to holding out one gold annotation and evaluating it against the other annotations, as done in prior work, which underestimates human performance relative to systems.}
For example, BERT, the current state-of-the-art on \squad{}, \emph{drops} by more than 50 absolute F1 points. This is a positive indication that \drop{} is indeed a challenging reading comprehension dataset, which opens the door for tackling new and complex reasoning problems on a large scale.

The best performance is obtained by our NAQANet model. \tabref{performance-by-type} shows that our gains are obtained on the challenging and frequent number answer type, which requires various complex types of reasoning. 
Future work may also try combining our model with BERT.
Furthermore, we find that all heuristic baselines do poorly on our data, hopefully attesting to relatively small biases in \drop.



\paragraph{Difficulties of building semantic parsers} We see that all the semantic parsing baselines perform quite poorly on \drop. This is mainly because of our pipeline of extracting tabular information from paragraphs, followed by the denotation-driven logical form search, can yield logical forms only for a subset of the training data. For SRL and syntactic dependency sentence representation schemes, the search was able to yield logical forms for 34\% of the training data, whereas with OpenIE, it was only 25\%. On closer examination of a sample of 60 questions and the information extracted by the SRL scheme (the best performing of the three), we found that only 25\% of the resulting tables contained information needed to the answer the questions. These observations show that high quality information extraction is a strong prerequisite for building semantic parsers for \drop. Additionally, the fact that this is a weakly supervised semantic parsing problem also makes training hard. The biggest challenge in this setup is the spuriousness of logical forms used for training, where the logical form evaluates to the correct denotation but does not actually reflect the semantics of the question. This makes it hard for the model trained on these spurious logical forms to generalize to unseen data. From the set of logical forms for a sample of 60 questions analyzed, we found that only 8 questions (13\%) contained non-spurious logical forms.


\paragraph{Error Analysis}

Finally, in order to better understand the outstanding challenges in \drop, we conducted an error analysis on a random sample of 100 erroneous NAQANet predictions.
The most common errors were on questions which 
required complex type of reasoning, such as arithmetic operations (evident in 51\% of the errors), counting (30\%), domain knowledge and common sense (23\%), co-reference (6\%), or a combination of different types of reasoning (40\%). See  \tabref{error_analysis} for examples of some of the common phenomena.


\begin{table}
    \centering
    \small
    \resizebox{\columnwidth}{!}{
    \begin{tabular}{lccccc}
    \toprule
    \multirow{2}{*}{\bf Type} & \multirow{2}{*}{$(\%)$} &\multicolumn{2}{c}{\bf Exact Match} & \multicolumn{2}{c}{\bf F1} \\
    \cmidrule(lr){3-4}
    \cmidrule(lr){5-6}
    & & QN+ & BERT & QN+ &  BERT \\
    \midrule
    Date        & 1.57  & 28.7   & 38.7 & 35.5  & 42.8\\
    Numbers     & 61.94 & 44.0  & 14.5 & 44.2  & 14.8 \\
    Single Span & 31.71 & 58.2 & 64.6  & 64.6  & 70.1 \\
    $>1$ Spans  & 4.77  & 0     & 0 & 17.13  & 25.0 \\
    \bottomrule
    \end{tabular}}
    \caption{Dev set performance breakdown by different answer types; our model (NAQANet, marked as \emph{QN+}) vs. BERT, the best-performing baseline.}
    \label{tab:performance-by-type}
\end{table}

\section{Conclusion}

We have presented DROP, a dataset of complex reading comprehension questions that require \textbf{D}iscrete \textbf{R}easoning \textbf{O}ver \textbf{P}aragraphs.  This dataset is substantially more challenging than existing datasets, with the best baseline achieving only 32.7\% F1, while humans achieve 96\%.  We hope this dataset will spur research into more comprehensive analysis of paragraphs, and into methods that combine distributed representations with symbolic reasoning.  We have additionally presented initial work in this direction, with a model that augments QANet with limited numerical reasoning capability, achieving 47\% F1 on DROP.

\section*{Acknowledgments}
We would like to thank Noah Smith, Yoav Goldberg, and Jonathan Berant for insightful discussions that informed the direction of this work.
The computations on \url{beaker.org} were supported in part by credits from Google Cloud.

%

\bibliography{paper}
\bibliographystyle{acl_natbib}

\newpage
\appendix
\begin{figure*}[t]
    \centering
    \begin{subfigure}[t]{\textwidth}
        \centering
        \includegraphics[width=1.0\textwidth]{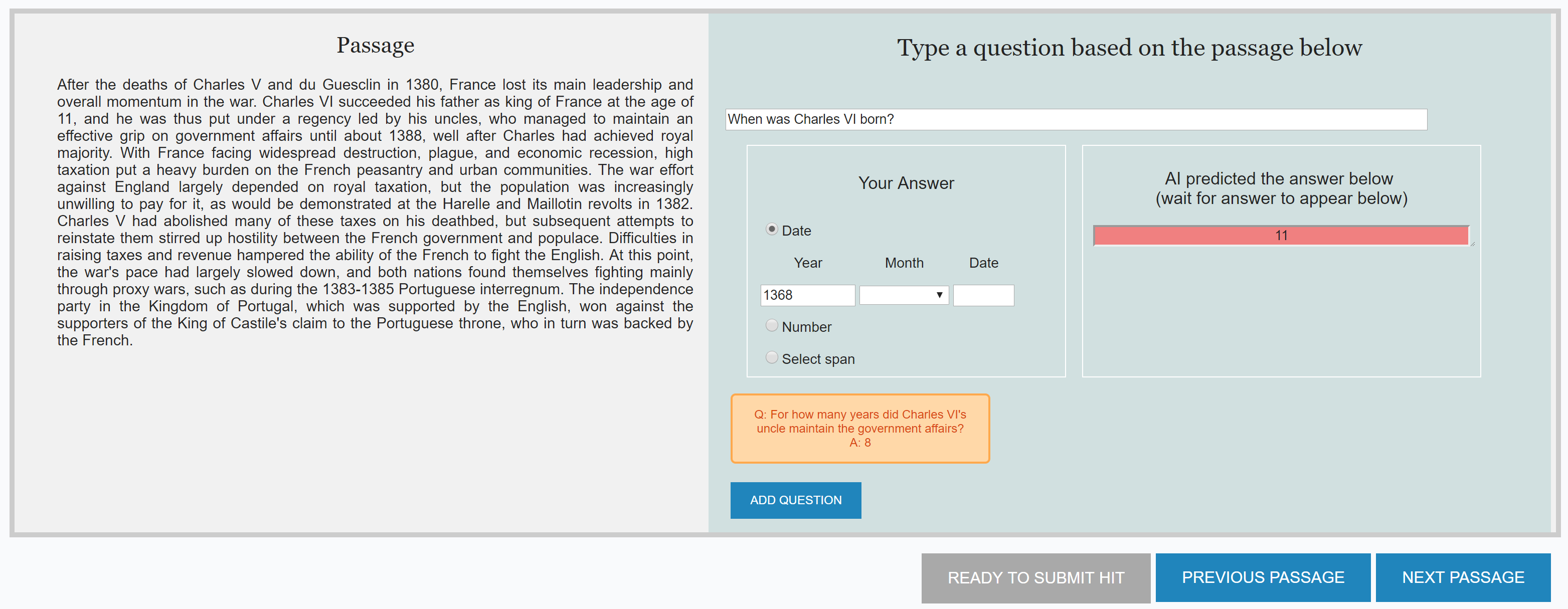}
    \end{subfigure}
    \begin{subfigure}[t]{\textwidth}
        \centering
        \includegraphics[width=0.75\textwidth]{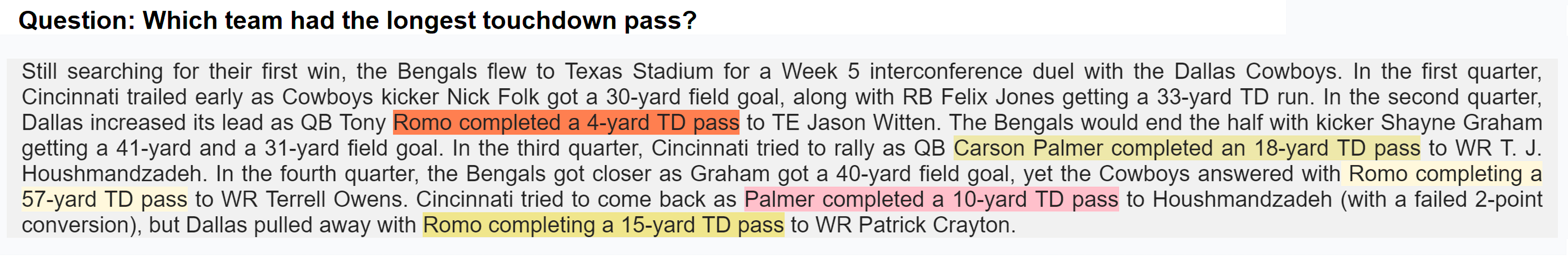}
    \end{subfigure}
    \begin{subfigure}[t]{\textwidth}
        \centering
        \includegraphics[width=0.75\textwidth]{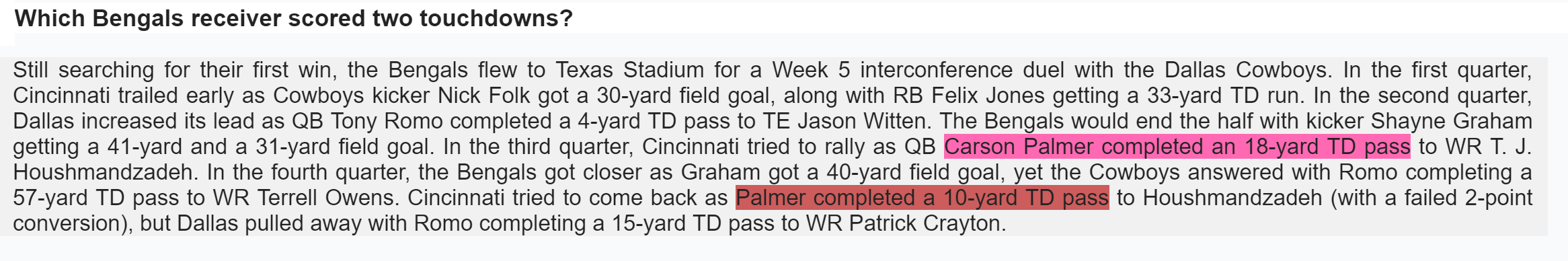}
    \end{subfigure}
     \begin{subfigure}[t]{\textwidth}
        \centering
        \includegraphics[width=0.75\textwidth]{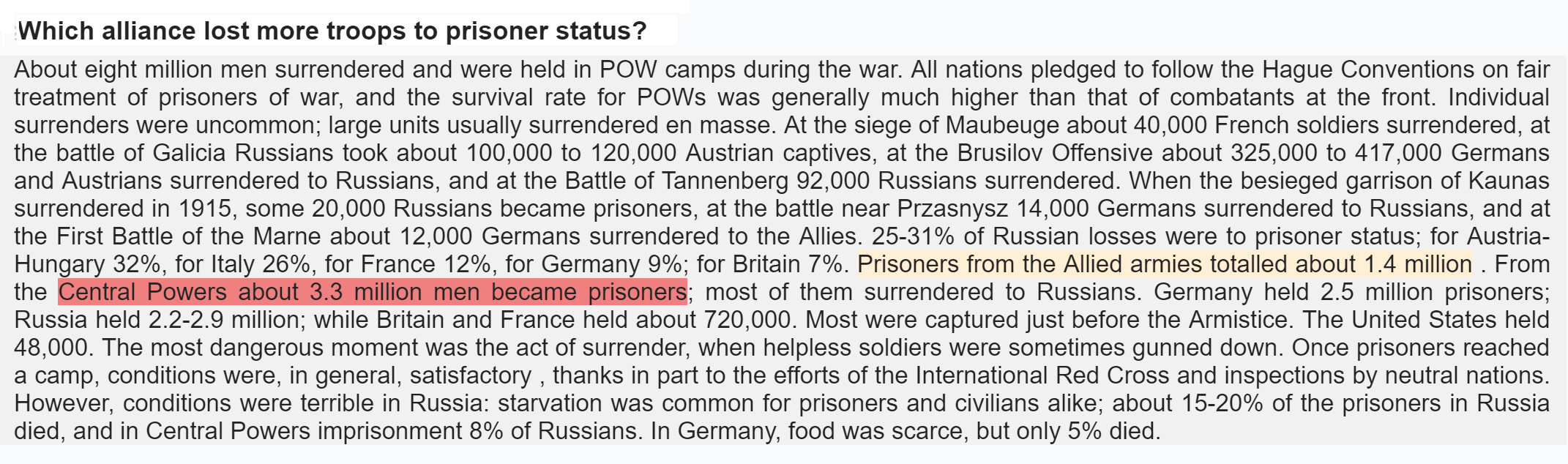}
    \end{subfigure}
     \begin{subfigure}[t]{\textwidth}
        \centering
        \includegraphics[width=0.75\textwidth]{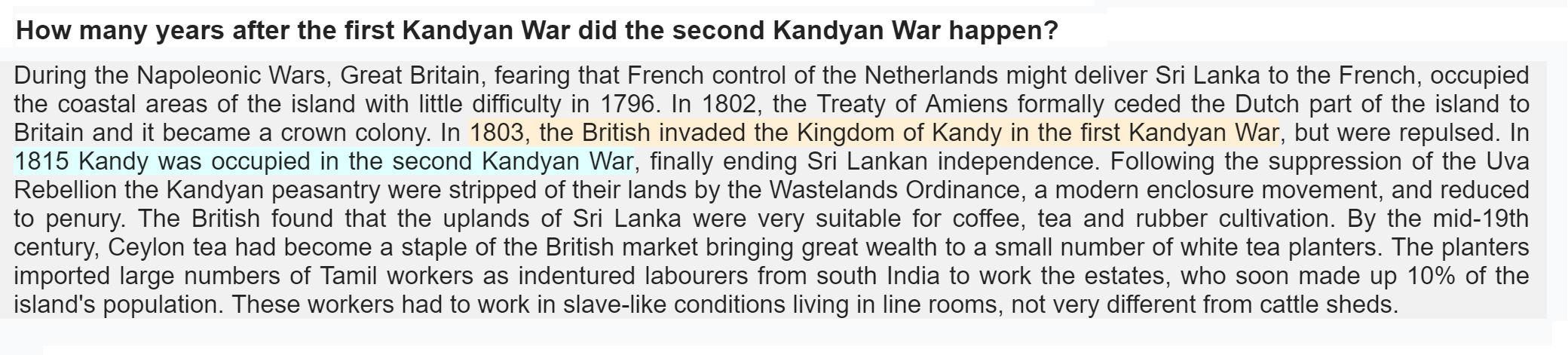}
    \end{subfigure}

    \caption{Question Answering HIT sample above with passage on the left and input fields for answer on the right and Highlighted candidate spans of sample answers below}
    \label{fig:question_types}
\end{figure*}

    



\end{document}